%% file: main.tex
\documentclass[10pt,twocolumn,letterpaper]{article}

\usepackage{cvpr}              
\input{preamble}
\definecolor{cvprblue}{rgb}{0.21,0.49,0.74}
\usepackage[pagebackref,breaklinks,colorlinks,allcolors=cvprblue]{hyperref}
\usepackage{url}
\usepackage{algorithm}
\usepackage{algorithmic}
\usepackage{booktabs}
\usepackage{graphicx}
\usepackage[dvipsnames]{xcolor}
\usepackage{xspace}
\usepackage{multirow}
\usepackage{pifont}
\usepackage{amssymb}
\usepackage{bbding}
\usepackage[skip=0pt]{caption}
\usepackage{siunitx}
\def\paperID{4802} 
\def\confName{CVPR}
\def\confYear{2026}

\title{Proxy-GS: Unified Occlusion Priors for Training and Inference in Structured 3D Gaussian Splatting}

\author{
\textbf{Yuanyuan Gao}$^{1,2}$\footnotemark[1]\quad
\textbf{Yuning Gong}$^{1,3}$\footnotemark[1]\quad
\textbf{Yifei Liu}$^{1}$\quad
\textbf{Jingfeng Li}$^{2}$\quad
\textbf{Dan Xu}$^{4}$\\
\textbf{Yanci Zhang}$^{3}$\quad
\textbf{Dingwen Zhang}$^{2}$\footnotemark[2]\quad
\textbf{Xiao Sun}$^{1}$\quad
\textbf{Zhihang Zhong}$^{5,1}$\footnotemark[2]\\
$^{1}$Shanghai Artificial Intelligence Laboratory \\
$^{2}$Northwestern Polytechnical University \\
$^{3}$Sichuan University \\
$^{4}$Hong Kong University of Science and Technology \\
$^{5}$Shanghai Jiao Tong University \\
}

\begin{document}


\twocolumn[{
\maketitle
    \begin{center}
        \centering
        \includegraphics[width= 1.0\linewidth ]{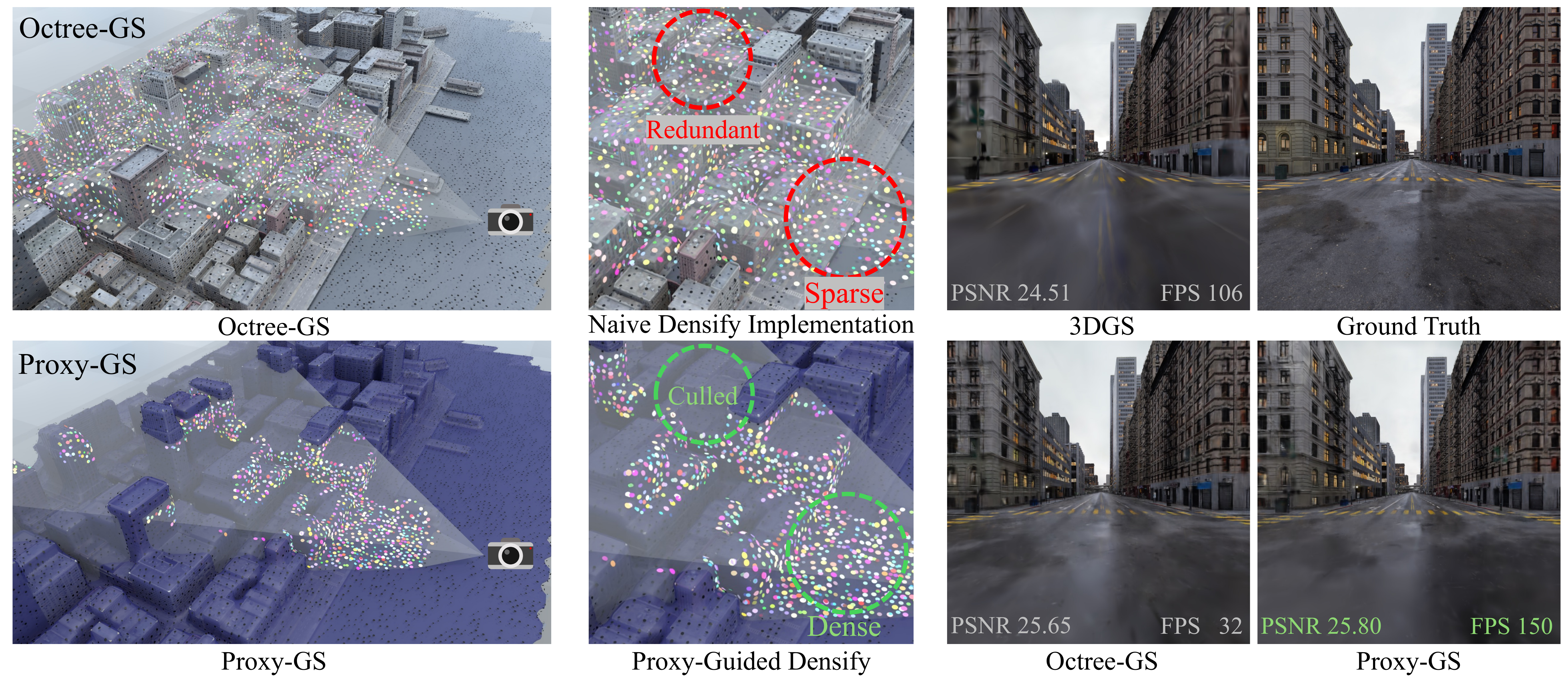}
        \captionof{figure}{
            We propose \textbf{Proxy-GS}, an occlusion-aware training and inference framework built upon lightweight proxies. By introducing proxy-guided densification, our method effectively guides anchors to grow in more geometrically meaningful regions. As a result, Proxy-GS not only achieves higher rendering quality but also delivers significantly faster rendering compared to state-of-the-art MLP-based 3DGS approaches.
        }
        \label{fig:teaser}
    \end{center}%
}]

\renewcommand{\thefootnote}{\fnsymbol{footnote}}


\footnotetext[1]{Denotes equal contribution. This work was done during their internship at Shanghai Artificial Intelligence Laboratory.}
\footnotetext[2]{Denotes corresponding author.}

\input{sec/0_abstract}    
\input{sec/1_intro}
\input{sec/2_relatedwork}

\input{sec/3_preliminaries}
\input{sec/4_method}

\input{sec/5_experiment}

\input{sec/6_conclusion}

\input{sec/7_acnowledgement}
{
    \small
    \bibliographystyle{ieeenat_fullname}
    \bibliography{main}
}

\input{sec/X_suppl}


\end{document}

%% file: sec/0_abstract.tex
\begin{abstract}
3D Gaussian Splatting (3DGS) has emerged as an efficient approach for achieving photorealistic rendering. Recent MLP-based variants further improve visual fidelity but introduce substantial decoding overhead during rendering. To alleviate computation cost, several pruning strategies and level-of-detail (LOD) techniques have been introduced, aiming to effectively reduce the number of Gaussian primitives in large-scale scenes. However, our analysis reveals that significant redundancy still remains due to the lack of occlusion awareness. In this work, we propose Proxy-GS, a novel pipeline that exploits a proxy to introduce Gaussian occlusion awareness from any view.
At the core of our approach is a fast proxy system capable of producing precise occlusion depth maps at resolution 1000$\times$1000 under \SI{1}{ms}. This proxy serves two roles: first, it guides the culling of anchors and Gaussians to accelerate rendering speed. Second, it guides the densification towards surfaces during training, avoiding inconsistencies in occluded regions, thus improving the rendering quality. 
In heavily occluded scenarios such as the MatrixCity Streets dataset, Proxy-GS achieves more than $2.5\times$ speedup over Octree-GS while also improving rendering quality.
\end{abstract}

%% file: sec/1_intro.tex
\section{Introduction}
With the emergence of Neural Radiance Fields (NeRF)~\citep{mildenhall2020nerf}, high-quality novel view synthesis has become possible, but the slow rendering speed limits its practical use. Recently, 3D Gaussian Splatting (3DGS)~\citep{kerbl3Dgaussians} has significantly improved efficiency, greatly advancing AR and VR applications, and even extending to areas such as virtual   
Avatars~\citep{hu2024gauhuman,xu2025sequential,zhan2025towards,niu2025anicrafter} and spatial intelligence~\cite{SpatialLM, gao2026holi}. However, vanilla 3DGS often produces heavily redundant Gaussians that attempt to fit every training view while neglecting the underlying scene geometry. To address this limitation and pursue higher-fidelity representations, structured MLP-based Gaussian approaches such as scaffold-GS~\citep{lu2024scaffold} and Octree-GS~\citep{ren2024octree} have been introduced.

At the core of the MLP-based 3DGS method lies an MLP decoder that conditions on the camera viewing direction to dynamically generate Gaussian attributes. Although these structured Gaussian variants substantially strengthen the modeling of challenging and view-dependent details, they also impose extra decoding operations at inference, leading to increased computational cost. This drawback becomes particularly critical in large-scale scene reconstruction~\citep{liu2024citygaussian,lin2024vastgaussian,gao2025citygs,li2024dgtr,gao2024cosurfgs}, where the number of Gaussian primitives and rendering complexity grow dramatically, making efficient decoding and rendering indispensable. 

Although pruning strategies~\citep{fan2024lightgaussian,lee2024compact,liu2025maskgaussian} can be introduced to reduce redundancy, they inevitably lead to a loss in rendering quality. Meanwhile, following works~\citep{ren2024octree, kerbl2024hierarchical, cui2024letsgo}  employ a level-of-detail (LOD) structure to mitigate redundancies from distant scene contents, but this approach is has no address in an occlusion environment. In contrast, real-world scenarios are full of occlusions, especially in large-scale modern city streets and complex indoor environments with multiple rooms. For future ultra-large VR walkthroughs that seamlessly span from indoor to outdoor scenes, effective occlusion culling becomes an essential and intuitive requirement. 

Moreover, since most practitioners rely on consumer-grade GPUs rather than datacenter-oriented ones such as A100s, it is important to consider the hardware characteristics of these devices. Consumer GPUs, typically designed for gaming and graphics applications, are equipped with dedicated hardware rasterization units. The widespread adoption of 3DGS thus requires careful adaptation to hardware rasterization~\citep{fast_gauss}.

To address the above limitations, we propose \textbf{Proxy-GS}, a proxy-guided Gaussian representation that leverages lightweight proxy meshes obtained through dedicated design.
By bridging hardware rasterization with a PyTorch-based proxy renderer, Proxy-GS can efficiently cull occluded anchors with negligible time consumption and seamlessly integrate this process with the original frustum selection strategy.
Furthermore, during training, the proxy guidance is incorporated again to provide stronger structural cues for anchor selection and densification.

As shown in Fig.~\ref{fig:teaser}, Proxy-GS not only achieves up to a $3\times$ speedup in rendering on top of existing MLP-based LOD frameworks Octree-GS~\citep{ren2024octree} but also improves occlusion awareness in anchor selection, leading to higher rendering quality.
Our main contributions can be summarized as follows:
\begin{itemize}
    \item We design a \textbf{proxy-guided training pipeline} that incorporates structural priors from proxy meshes, enabling MLP-based approaches to be occlusion-aware and achieve higher rendering quality.
    \item Under a consistent training and testing setting, Proxy-GS achieves more than a $3\times$ FPS speedup over the LOD baseline on occlusion-rich scenes, while simultaneously improving rendering quality; in scenes with few occlusions, it introduces virtually no overhead.
    \item We leverage hardware rasterization to reduce the time of acquiring a $1000^2$-resolution depth map to under \SI{1}{ms}.
\end{itemize}

%% file: sec/2_relatedwork.tex
\section{Related Work}
\subsection{Neural Rendering}
Neural Radiance Fields (NeRFs)~\citep{mildenhall2020nerf} pioneered the idea of representing a scene as a volumetric radiance field, enabling high-quality novel view synthesis for bounded scenes, typically centered around a single object. Subsequent extensions improved the scalability and visual fidelity of NeRF-based methods: Mip-NeRF~\citep{barron2021mip} introduced proper anti-aliasing to handle multi-scale observations, NeRF++~\citep{zhang2020nerf++} lifted the constraint of strictly bounded scenes, and Mip-NeRF 360~\citep{mipnerf360} extended anti-aliased representations to unbounded, object-centric settings. 
Despite these advances, NeRF-style volumetric rendering remains computationally expensive due to the need for dense ray sampling and neural field evaluation. 

To overcome this inefficiency, 3D Gaussian Splatting (3DGS)~\citep{kerbl20233d} was recently proposed as an explicit point-based alternative. 
While 3DGS achieves real-time performance with explicit Gaussian primitives, its reliance on directly optimized parameters often leads to limited expressiveness, particularly in capturing fine-grained appearance details and complex view-dependent effects. 
To address this shortcoming, MLP-based extensions such as Scaffold-GS~\citep{lu2024scaffold} and Octree-GS~\citep{ren2024octree} introduce neural decoders that generate Gaussian attributes from learned anchor features. By leveraging structured anchors and neural decoding, these approaches significantly improve the representational capacity, enabling more accurate modeling of geometry and appearance in large and challenging scenes.  


\subsection{Efficient 3D Gaussian Splatting Rendering}
For rendering acceleration, many studies~\citep{lee2024compact, fan2024lightgaussian, li2024dgtr, wang2025hyrf} have explored pruning or compression strategies to reduce the number of Gaussians and thus alleviate computational overhead. While such pruning-based methods can be effective to some extent, they inevitably face scalability bottlenecks in large scenes, where aggressive pruning results in performance degradation. 
Beyond the pruning-based strategy, another line of research focuses on architectural designs for rendering acceleration. Among them, level-of-detail (LOD) architectures have become particularly influential. Hierarchical-GS~\citep{kerbl2024hierarchical} merges neighboring Gaussians to reduce rendering cost, achieving higher frame rates at the expense of some visual fidelity. LetsGo~\citep{cui2024letsgo} jointly optimizes multi-resolution Gaussian models and demonstrates strong performance in LiDAR-based scenarios, yet its reliance on multi-resolution point cloud inputs incurs substantial training overhead and creates a strong dependence on point cloud accuracy. CityGaussian~\citep{liu2024citygaussian} further combines pruning strategies~\citep{fan2024lightgaussian} with LOD-based rendering to enhance scalability in urban scenes. {LODGE}~\citep{kulhanek2025lodge} proposes an LOD framework that adapts visible Gaussians across scales for memory and speed-constrained rendering. {Horizon-GS}~\citep{jiang2025horizon} unifies aerial-to-ground reconstruction and rendering with scalable data organization for large environments. {Virtualized 3D Gaussians}~\citep{yang2025virtualized} introduces a cluster-based LOD with hierarchical Gaussian groups and online selection for composed scenes, and {VastGaussian}~\citep{lin2024vastgaussian} demonstrates city-scale reconstruction via progressive partition and merging. 

While the aforementioned works improve efficiency for explicit 3DGS, more LOD mechanisms have also been extended to MLP-based Gaussians. Octree-GS~\citep{ren2024octree} organizes anchors into a multi-level octree, where the level selection is determined by the distance to the camera, thereby reducing the number of anchors decoded at each frame. This strategy alleviates part of the computational burden in large-scale scenes, but the rendering speed still leaves considerable room for improvement. Recent work Cache-GS~\citep{Tao2025GSCacheAG} provides acceleration by reusing decoded Gaussians, doubling the rendering speed of Octree-GS, although this comes with a noticeable loss in rendering quality. 
In parallel, methods like FLASH-GS~\citep{Feng2024FlashGSE3} target low-level CUDA optimizations of the original 3DGS pipeline, aiming to improve efficiency at the kernel level. There has also been progress in web-based Gaussian splatting rendering. For example, works like Visionary~\citep{gong2025visionary} utilize WebGPU for efficient Gaussian sorting, thereby accelerating inference and visualization directly within browsers.
Aside from the low-level optimizations, OccluGaussian~\citep{liu2025occlugaussian} reasons about occlusion by partitioning scenes into clusters, our approach performs per-pixel, proxy-guided filtering, which preserves fine details and aligns with actual rendering cost.
Recent work has also explored leveraging occlusion for accelerating rendering. For example, ~\citet{Ye2025WhenGM} proposed using pre-rendered depth maps to guide 3DGS rendering. However, their depth acquisition relies on surfel rendering, which is less efficient compared to our lightweight proxy-based approach.

%% file: sec/3_preliminaries.tex
\section{Preliminaries}
\subsection{MLP-based 3DGS} 
To exploit the structural priors provided by Structure-from-Motion (SfM), a line of work such as Scaffold-GS~\citep{lu2024scaffold} and Octree-GS~\citep{ren2024octree} has been developed. Instead of reconstructing Gaussians directly from sparse SfM points, Scaffold-GS first builds a coarse voxel grid and places anchor points at the voxel centers. Each anchor is associated with a latent feature vector $f$, which is fed into a multi-layer perceptron (MLP) to decode the corresponding Gaussian attributes:
\begin{equation}
\{ \mu_j, \Sigma_j, c_j, \alpha_j \}_{j \in \mathcal{M}} 
= \mathrm{MLP}_\theta(f_i, v_i)_{i \in \mathcal{N}},
\label{eq:anchor_mlp}
\end{equation}

where $\theta$ denotes the MLP parameters, and $\mu_j$, $\Sigma_j$, $c_j$, and $\alpha_j$ represent the mean, covariance, color, and opacity of the $j$-th Gaussian derived from the $i$-th anchor under viewing direction $v_i$. The generated neural Gaussians are subsequently rasterized in the same way as explicit 3D Gaussians.
The advantage of anchor-based placement is that the decoded Gaussians inherit structural cues from the underlying SfM prior, which reduces redundancy and improves robustness for novel view rendering. Octree-GS extends this framework by substituting the voxel grid with an explicit octree representation, enabling the scene to be modeled at multiple resolutions.
The hierarchical design of the octree naturally supports level-of-detail (LOD) construction. During rendering, appropriate LOD levels can be selected adaptively based on the camera distance, thereby reducing decoding cost and improving scalability to larger-scale scenes.

\subsection{Hardware Rasterization} 
Hardware rasterization denotes the GPU’s fixed/near–fixed-function graphics path that transforms vertices to clip/NDC, discretizes primitives into fragments, interpolates attributes, and resolves visibility via depth/stencil tests and blending before writing to render targets. This behavior is standardized in modern graphics APIs and is executed by specialized units. The pixel backend, commonly called the Raster Operations Processor (ROP, a.k.a. render output unit) houses depth/stencil units that perform depth and stencil tests and update the corresponding buffers, and color units that handle blending, format conversion/MSAA resolves, and render-target writes. 
These mechanisms underpin the extreme throughput and bandwidth efficiency of the pipeline. Architecturally, rasterization evolved from fixed-function to programmable/unified shader models and is realized across immediate-mode and tile/binning GPU designs, but the visibility tests and depth buffering remain conceptually consistent. In this work, we will later exploit this machinery in a depth-only pass on a proxy mesh to obtain a conservative Z-buffer at negligible cost, which we then consume as a visibility prior.

\begin{figure*}[!t]
    \centering
    \includegraphics[width=1\linewidth]{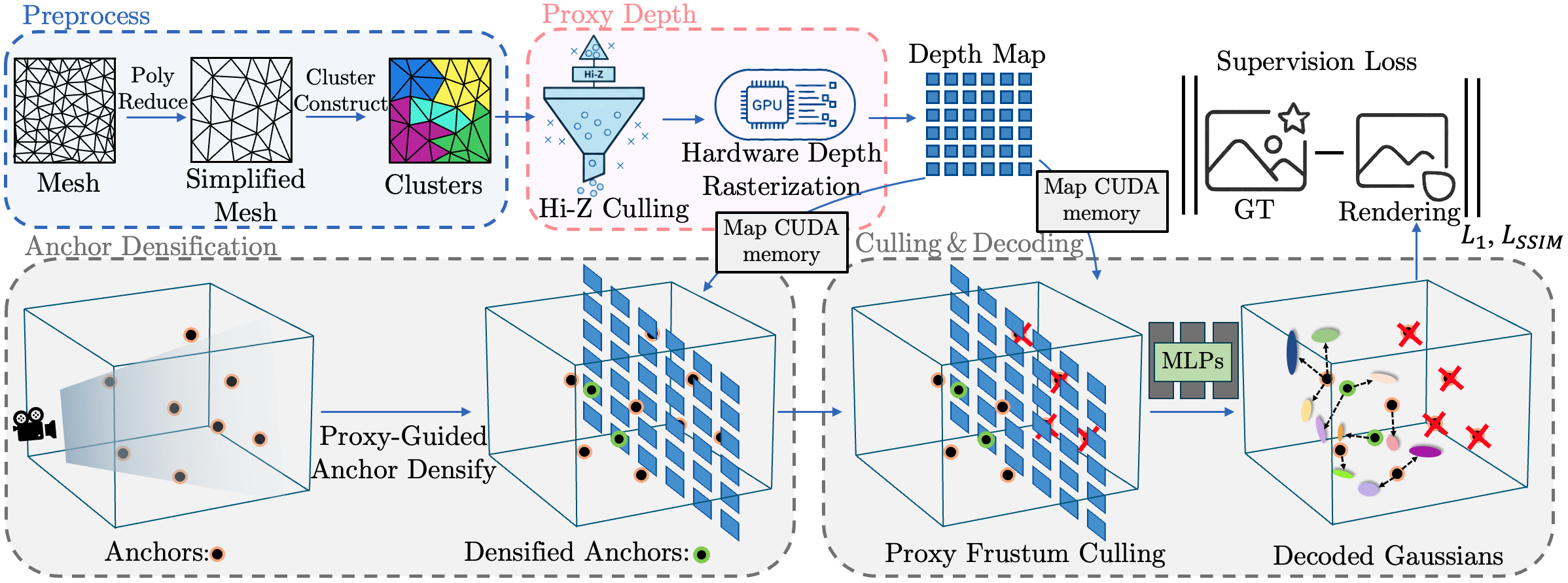}
    \vspace{1pt}
    \caption{
    \textbf{Proxy-GS Framework.} 
    We first construct a lightweight proxy mesh. During rendering, hardware rasterization produces a depth map in under 1\,ms, which is then used to efficiently cull anchors that are occluded. During training, in addition to the same rendering pipeline, we further introduce structure-aware anchor densification, encouraging anchors to grow adaptively along the proxy mesh geometry.
    }
    \label{fig:framework}
\vspace{-0.1in}
\end{figure*}

%% file: sec/4_method.tex
\section{Method}
\subsection{Motivation}
Reconstructing large-scale scenes with high occlusion presents unique challenges due to the vast number of Gaussians and anchors involved. As illustrated in Fig.~\ref{fig:teaser},
When visualizing the anchors used for decoding, we observe a significant mismatch between the decoded anchors and those that are intuitively required for accurate rendering. In particular, a large proportion of anchors correspond to heavily occluded regions, which substantially increases the decoding burden without contributing to the final image quality. Effective occlusion culling, therefore, has the potential to greatly reduce computational cost.  

Existing MLP-based works, such as Octree-GS~\cite{ren2024octree} and Scaffold-GS~\cite{lu2024scaffold}, design anchor structures to better exploit the inherent hierarchy and structural priors. However, since their anchor selection does not explicitly account for occlusions, the anchors are optimized merely to fit RGB images. As a result, the binding between anchors and their associated Gaussians can become inconsistent in space, leading to redundant decoding and degraded structural interpretability.

\subsection{Proxy Guided Filter}
A central question in our study is how to obtain occlusion relationships both efficiently and with negligible loss of accuracy. We find that leveraging lightweight proxy meshes for hardware rasterization enables depth rendering at only a marginal time cost. For many outdoor large-scale scenes, dense point clouds are already available or can be generated using tools such as COLMAP. In contrast, indoor scenes often contain texture-less regions that cause SfM-based reconstruction to fail. As large reconstruction models~\cite{wang2025vggt} already achieve very strong performance, especially for indoor scenes. In our pipeline, we leverage MapAnything~\cite{keetha2025mapanything}, using Colmap pose and RGB image as input to obtain a dense point cloud for indoor environments and then convert it into a mesh. Further implementation details are provided in the Appendix~\ref{mesh_extract}.
We construct proxy meshes using existing different pipelines and apply surface simplification to retain only coarse geometric structures. This proxy is sufficient to fully exploit the high throughput of hardware fixed-function units for efficient depth generation.

\begin{figure}[t]
    \centering
    \includegraphics[width=0.8\linewidth]{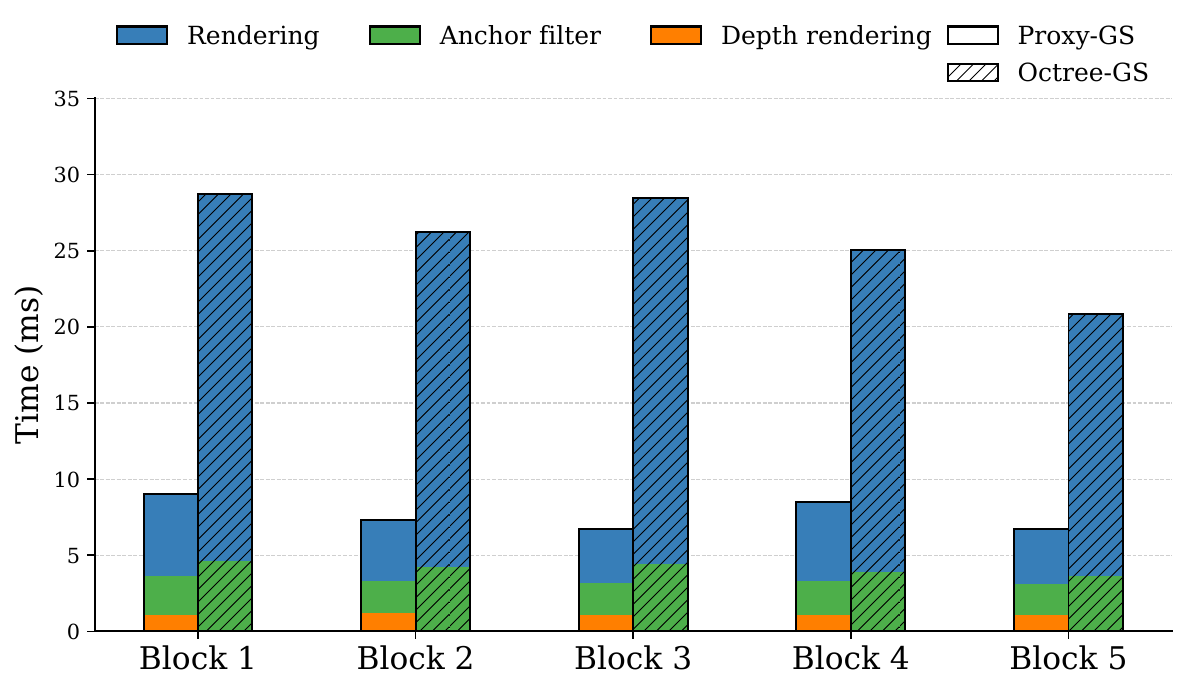}
    \captionsetup{skip=2pt} 
    \caption{\textbf{Comparison of the time proportion of each inference component (Rendering, anchor filter, depth rendering) with that of Octree-GS on MatrixCity dataset.} }
    \label{fig:time_comparison}
\vspace{-6pt}
\end{figure}

As shown in Fig. ~\ref{fig:framework}, to further accelerate the process, the mesh is partitioned into fine-grained clusters, and hierarchical visibility checks such as Hierarchical Z-buffer (Hi-Z) culling \cite{greene1993hierarchical} are employed to quickly cull invisible clusters. In the fragment stage, Early-Z is enabled, and we keep the fragment shader minimal by removing operations unrelated to depth writes. This allows our method to output depth maps at a high speed even in complex and large-scale urban scenes, as shown in Fig.~\ref{fig:time_comparison}. The depth map is kept on GPU and directly exploited in CUDA occlusion culling to avoid GPU-CPU-GPU round-trip overhead. We also compare the speed of our method with other conventional depth-acquisition approaches in the Appendix ~\ref{depth-acquisition}, further demonstrating our efficiency advantages.

Then we fuse the occlusion culling and frustum culling of anchors in a single CUDA kernel:
We denote the pixels ndc coordinates as $(x_{\text{ndc}},y_{\text{ndc}},z_{\text{ndc}})$.

A visibility check is then performed: points with  
$
z_{\text{h}} \leq \tau,
\quad \tau = 10^{-4},
$
are regarded as invalid (filtered), since they lie behind the camera or are too close to the near plane.
After projecting to normalized device coordinates (NDC), we map the coordinates to discrete pixel indices $(u,v)$:  
\begin{equation}
x_{\text{pix}} = \left\lfloor \frac{(x_{\text{ndc}}+1)}{2} \cdot W \right\rfloor, 
\quad 
y_{\text{pix}} = \left\lfloor \frac{(y_{\text{ndc}}+1)}{2} \cdot H \right\rfloor,
\end{equation}

where $W$ and $H$ denote the image width and height, respectively. 
A pixel is discarded if it falls outside the image boundary:  
\begin{equation}
x_{\text{pix}} < 0 \;\lor\; x_{\text{pix}} \geq W \;\lor\; y_{\text{pix}} < 0 \;\lor\; y_{\text{pix}} \geq H.
\end{equation}


For valid pixels, we retrieve the hardware depth $z_{hw}\in[0,1]$ at $(x_{\text{pix}},y_{\text{pix}})$ from the depth image.
We then convert it to the \emph{linear} camera-space depth using the near/far planes $n,f$:
\begin{equation}
d_{\text{mesh}}(x_{\text{pix}},y_{\text{pix}})
= \frac{n\,f}{\,f - z_{hw}(x_{\text{pix}},y_{\text{pix}})\,\big(f-n\big)\,}.
\end{equation}
Finally, we apply a small safety margin $\gamma$:
\begin{equation}
\hat{d}(x_{\text{pix}},y_{\text{pix}}) = d_{\text{mesh}}(x_{\text{pix}},y_{\text{pix}}) + \gamma .
\label{eq:depth_offset}
\end{equation}

If the depth value is invalid (e.g., in sky regions), the point is not culled. Otherwise, we apply the depth test:
\begin{equation}
\text{Cull}(\mathbf{p}) = 
\begin{cases}
\text{true}, & z_{\text{h}} > \hat{d}(x_{\text{pix}},y_{\text{pix}}), \\
\text{false}, & z_{\text{h}} \leq \hat{d}(x_{\text{pix}},y_{\text{pix}}).
\end{cases}
\end{equation}
\noindent To summarize, a point is removed if its camera-space depth lies \emph{behind} the depth map at the corresponding pixel, which effectively performs occlusion culling on the image plane.

\subsection{Proxy-Guided Densification}
In the original anchor-growing densification strategy, new anchors are generated around Gaussian splats that exhibit large gradients during training. However, this procedure may introduce redundant anchors behind the proxy mesh depth: although these Gaussians have large gradients, the new anchors feature do not decode due to occlusion.  

To tackle this limitation, and inspired by the multi-view depth densification strategy in Mvg-splatting~\cite{Li2024MVGSplattingMG}, we introduce \textbf{proxy-guided densification}, which explicitly projects anchors onto the surface of the proxy mesh. Since proxy depth maps are pre-computed, we can measure the patch-wise L1 loss and identify regions where the rendering error is consistently large. 


\begin{table*}[t]
\centering
\caption{Quantitative results on MatrixCity~\citep{li2023matrixcity}. 
We report average results over Block 1\&2, Block 3\&4, and Block 5. 
(Block 1\&2 and 3\&4 represent the average evaluation metrics of their respective two blocks.) 
The \textbf{best} and \underline{second-best} are highlighted.}
\label{tab:avg_blocks}
\footnotesize
\setlength{\tabcolsep}{4pt}
\renewcommand{\arraystretch}{1.1}
\resizebox{\textwidth}{!}{
\begin{tabular}{l|cccc|cccc|cccc}
\toprule[1.2pt]
& \multicolumn{4}{c|}{Block 1\&2} & \multicolumn{4}{c|}{Block 3\&4} & \multicolumn{4}{c}{Block 5} \\
Methods & PSNR$\uparrow$ & SSIM$\uparrow$ & LPIPS$\downarrow$ & FPS$\uparrow$
        & PSNR$\uparrow$ & SSIM$\uparrow$ & LPIPS$\downarrow$ & FPS$\uparrow$
        & PSNR$\uparrow$ & SSIM$\uparrow$ & LPIPS$\downarrow$ & FPS$\uparrow$ \\
\midrule
3DGS~\citep{kerbl20233d} & 21.55 & 0.730 & 0.366 & \underline{115} & 20.78 & 0.739 & 0.372 & \underline{114}   & 20.70 & 0.697 & 0.425 & \underline{121} \\
Scaffold-GS~\citep{lu2024scaffold} & 21.44 & 0.721 & 0.375 & 81 & 20.56 & 0.727 & 0.376 & 66 & 20.56 & 0.693 & 0.426 & 71 \\
Hierarchical-GS~\citep{kerbl2024hierarchical} & 20.50 & 0.707 & 0.418 & 61 & 20.38 & 0.719 & 0.422 & 41 & 20.22 & 0.673 & 0.463 & 60 \\
Hierarchical-GS($\tau_1$) & 20.50 & 0.706 & 0.419 & 62 & 20.38 & 0.718 & 0.424 & 45 & 20.22 & 0.672 & 0.466 & 66 \\
Hierarchical-GS($\tau_2$) & 20.46 & 0.702 & 0.423 & 71 & 20.30 & 0.711 & 0.431 & 49 & 20.20 & 0.671 & 0.467 & 75 \\
Hierarchical-GS($\tau_3$) & 20.01 & 0.678 & 0.450 & 85 & 19.71 & 0.680 & 0.464 & 63 & 20.01 & 0.657 & 0.483 & 90 \\
Octree-GS~\citep{ren2024octree} & \underline{21.94} & \underline{0.737} & \underline{0.347} & 32   & \underline{20.95} & \underline{0.743} & \underline{0.354} & 30   & \underline{21.41} & \underline{0.731} & \underline{0.375} & 48 \\
Proxy-GS & \textbf{22.11} & \textbf{0.751} & \textbf{0.330} & \textbf{126} & \textbf{21.06} & \textbf{0.751} & \textbf{0.348} & \textbf{134} & \textbf{21.68} & \textbf{0.744} & \textbf{0.362} & \textbf{151} \\
\bottomrule[1.2pt]
\end{tabular}
}
\end{table*}

\begin{table*}[t]
\centering
\caption{Quantitative results on real world outdoor and indoor datasets~\citep{xiong2024gauu,kerbl2024hierarchical,barron2023zip}. The \textbf{best} and \underline{second-best} are highlighted. Small City has more severe occlusions, while Berlin and CUHK-LOWER have relative weaker occlusions.}
\label{tab:real_world}
\footnotesize
\setlength{\tabcolsep}{4pt}
\renewcommand{\arraystretch}{1.1}
\resizebox{\textwidth}{!}{
\begin{tabular}{l|cccc|cccc|cccc}
\toprule[1.2pt]
& \multicolumn{4}{c|}{Small City} & \multicolumn{4}{c|}{Berlin} & \multicolumn{4}{c}{CUHK-LOWER} \\
Methods & PSNR$\uparrow$ & SSIM$\uparrow$ & LPIPS$\downarrow$ & FPS$\uparrow$
        & PSNR$\uparrow$ & SSIM$\uparrow$ & LPIPS$\downarrow$ & FPS$\uparrow$
        & PSNR$\uparrow$ & SSIM$\uparrow$ & LPIPS$\downarrow$ & FPS$\uparrow$ \\
\midrule
3DGS~\citep{kerbl20233d} 
&22.90  &0.727  &0.372  &\underline{132}  
&27.79  &0.907  &0.223  &187    
&25.48  &0.729  &0.389  &138  \\

Scaffold-GS~\citep{lu2024scaffold} 
&20.00  &0.713  &0.370  &62 
&27.80  &\textbf{0.912}  &\textbf{0.213}  &128  
&26.30  &0.785  &0.282  &117 \\

Hierarchical-GS~\citep{kerbl2024hierarchical} 
&22.07  &0.728  &0.377  &89  
&27.65  &0.902 &0.228  &145  
&25.18 &0.707 &0.408 &90  \\

Hierarchical-GS($\tau_1$) 
&22.07  &0.728  &0.377  &90  
&27.65  &0.901 &0.229  &150    
&25.19  &0.708  &0.408  &82  \\

Hierarchical-GS($\tau_2$) 
&22.07  &0.728  &0.378  &106  
&27.60  &0.899  &0.232  &152    
&25.14  &0.705  &0.411  &96  \\

Hierarchical-GS($\tau_3$) 
&22.02  &0.722  &0.386  &119  
&27.34  &0.890  &0.244  &160   
&24.58  &0.678  &0.435  &120  \\

Octree-GS~\citep{ren2024octree} 
&\underline{23.03}  &\underline{0.731}  &\underline{0.355}  &51 
&\underline{27.83}  &0.911  &0.218  &\underline{263}   
&\underline{26.42}  &\underline{0.794}  &\underline{0.267}  &\underline{212} \\

Proxy-GS 
&\textbf{23.09}  &\textbf{0.736}  &\textbf{0.344}  &\textbf{139} 
&\textbf{27.85}  &\textbf{0.912}  &\underline{0.216}  &\textbf{275}  
&\textbf{26.44}  &\textbf{0.795}  &\textbf{0.262}  &\textbf{239} \\
\bottomrule[1.2pt]
\vspace{-2pt}
\end{tabular}
}
\end{table*}

To achieve this, patches with abnormally high error are identified by comparing to the mean error $\bar{\ell}$ within the same frame.
We compute the per-patch loss as the average of pixel losses:
\[
\ell_{\mathcal P}=\frac{1}{|\Omega_{\mathcal P}|}\!\sum_{(u,v)\in\Omega_{\mathcal P}}\ell(u,v),
\qquad
\bar{\ell}=\frac{1}{|\mathcal S|}\!\sum_{\mathcal P\in\mathcal S}\ell_{\mathcal P}.
\]
We select patches that satisfy
\[
\ell_{\mathcal P}>\tau,\qquad \tau=3\,\bar{\ell}.
\]
For each selected patch $\mathcal P$, we choose a pixel $(u_{\mathcal P},v_{\mathcal P})$ (e.g., the patch center),
read the depth obtained from hardware rasterization $z_h(u_{\mathcal P},v_{\mathcal P})$, and convert it to linear camera-space depth with near/far $(n,f)$, to obtain $d_{\text{mesh}}(u_{\mathcal P},v_{\mathcal P})$. 
We then back-project this pixel to 3D and take it as the new anchor position:
\[
\hat{\mathbf p}_{\mathcal P}
= \mathbf o + \mathbf R^\top\!\left(
d_{\text{mesh}}(u_{\mathcal P},v_{\mathcal P})\,
\mathbf K^{-1}\!\begin{bmatrix}u_{\mathcal P}\\[2pt] v_{\mathcal P}\\[2pt] 1\end{bmatrix}
\right),\qquad
\mathbf a \leftarrow \hat{\mathbf p}_{\mathcal P}.
\]


To prevent redundancy in 3D space, we maintain a proxy-grid with cell size $h$ and origin $\mathbf b_{\min}$,
and allow up to $K$ anchors per cell:
\[
\mathbf c(\mathbf a)=\left\lfloor\frac{\mathbf a-\mathbf b_{\min}}{h}\right\rfloor\in\mathbb Z^3,
\qquad
\text{insert } \mathbf a \ \text{if}\ \kappa\!\left[\mathbf c(\mathbf a)\right] < K,
\]
where $\kappa[\cdot]\in\mathbb{N}$ counts the anchors in each cell.

%% file: sec/5_experiment.tex
\section{Experiment}
\paragraph{Datasets.} We begin by comparing our approach with other methods on the large-scale urban dataset MatrixCity~\citep{li2023matrixcity} to assess rendering quality. 
We follow the partition script of the MatrixCity, and divided the 8477 street images in its Small City into 5 blocks. Details can be seen in the Appendix~\ref{partition}.
The evaluation is further extended to indoor scenes from Zip-NeRF~\citep{barron2023zip}. 
In addition, we also test on real-world scenes that have different levels of occlusion and scale, including a street scene: Small City dataset~\citep{kerbl2024hierarchical}, and aerial-view scenes from CUHK-LOWER~\citep{xiong2024gauu}. We select both indoor and aerial top-down scenes to examine whether our method incurs additional overhead in relatively small or minimally occluded environments.
\vspace{-6pt}
\noindent\paragraph{Evaluation Criterion.} 
We adopt three widely used image quality metrics to evaluate novel view synthesis: peak signal-to-noise ratio (PSNR), structural similarity index (SSIM), and learned perceptual image patch similarity (LPIPS)~\citep{zhang2018unreasonable}. 
In addition, we report frames per second (FPS) to measure the rendering speed of different methods.

\begin{figure*}[t]
    \centering
    \includegraphics[width=1\linewidth]{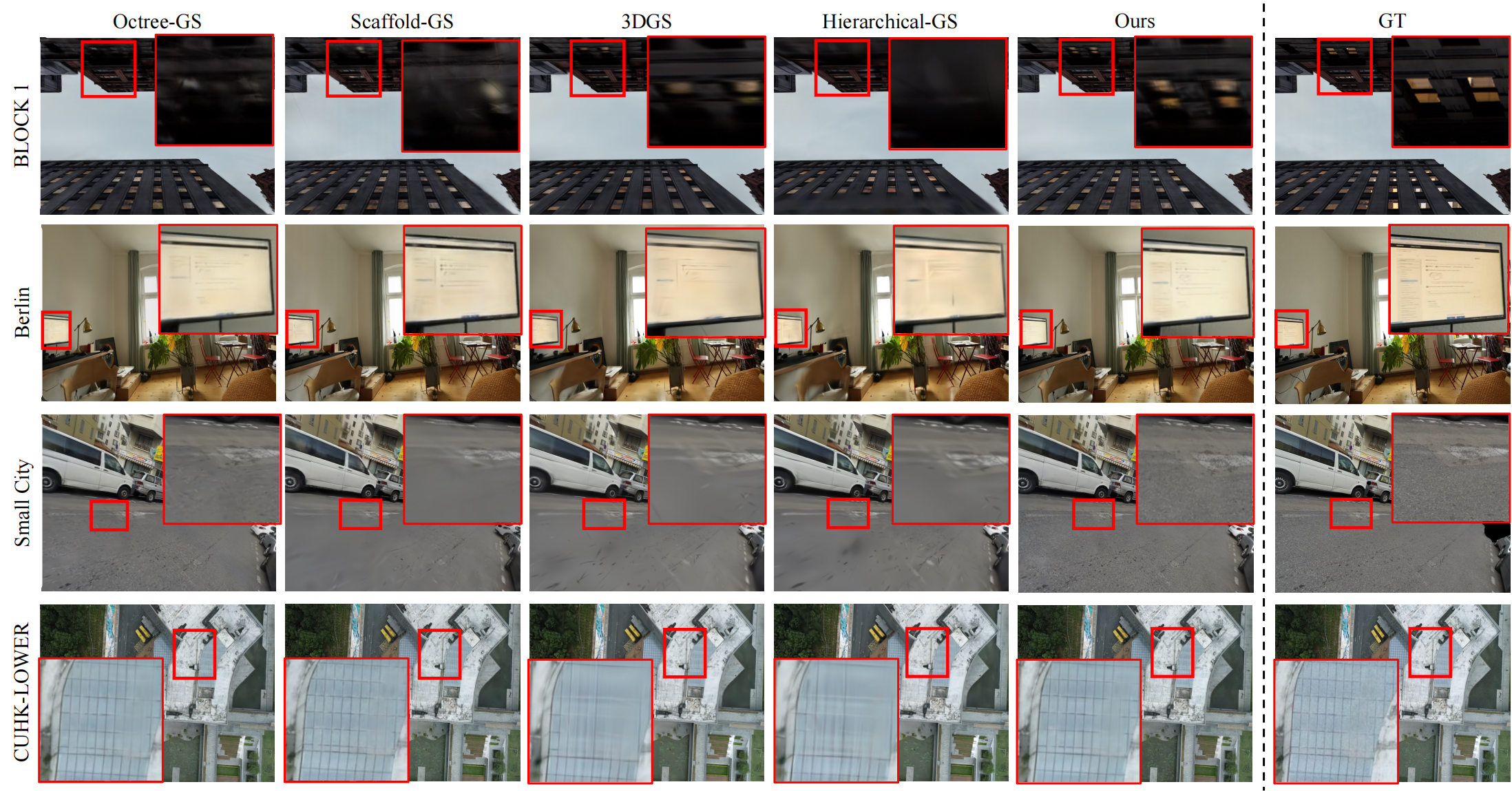}
    \caption{\textbf{Qualitative comparison.}
    Visualization on different datasets, where the regions with noticeable differences are highlighted and zoomed in with red boxes~\citep{li2023matrixcity, xiong2024gauu, barron2023zip, kerbl2024hierarchical}.
    }
    \label{fig:visualization}
\end{figure*}
\paragraph{Implementation Details.} 
Our method is implemented on top of the state-of-the-art MLP-based Octree-GS~\citep{ren2024octree}, following its default initialization and LOD strategy. 
For comparison, we also re-implement 3DGS~\citep{kerbl20233d}, Scaffold-GS~\citep{lu2024scaffold}, and Hierarchical-GS~\citep{kerbl2024hierarchical}, and train all methods for 40k iterations. Specifically, for the evaluation of Hierarchical-GS, we set the $\tau_1, \tau_2, \tau_3 = 3, 6, 15$. 
For approaches that do not employ MLPs, such as 3DGS and Hierarchical-GS, their default configurations typically yield higher rendering FPS but exhibit a noticeable quality gap compared to Octree-GS.
Since an increased number of Gaussian primitives generally leads to better rendering quality~\citep{zhao2024scaling}, we reduce the densification threshold to $10^{-4}$ across all scenes to ensure a fair comparison, resulting in rendering quality closer to that of Octree-GS. 
Unlike Octree-GS, Scaffold-GS initializes with fewer anchors due to the absence of multi-round sampling. To improve its rendering fidelity, we adopt a smaller voxel size of $10^{-4}$ together with a lower densification threshold of $10^{-4}$. 
All training experiments are performed on a single NVIDIA A100-40GB GPU. For inference, we employ a consumer-grade RTX 4090 GPU to reflect real-world deployment scenarios better.
\noindent\subsection{Main Results}
\paragraph{Novel View Synthesis and rendering FPS.}
As shown in Tab.~\ref{tab:avg_blocks} and Tab.~\ref{tab:real_world}, our method achieves higher or comparable rendering quality compared to all other baselines. Moreover, Fig.~\ref{fig:visualization} illustrates that our approach better preserves fine details such as building windows and crosswalk patterns.
In particular, as shown in Tab.~\ref{tab:avg_blocks}, the large urban street scenes simulated in MatrixCity are highly suited to our approach, where we consistently outperform existing methods in both rendering quality and speed.

Furthermore, to demonstrate the generality of our method, in Tab.~\ref{tab:real_world}
We also evaluate our method on aerial-view scenes, indoor environments, and real-world town streets, where it achieves comparable or superior performance against current state-of-the-art methods. Although aerial scenes typically involve limited occlusions and the current indoor dataset often contains relatively few rooms with sparse occlusion patterns, our method still yields noticeable improvements. Moreover, for small-city street scenes, which bear resemblance to the MatrixCity dataset, our approach delivers substantial improvements over the MLP-based method Octree-GS, achieving higher rendering quality while boosting FPS by $2.73\times$. These results collectively demonstrate the broad applicability of our method across diverse scenarios, while also highlighting that the extent of performance gains may vary depending on the characteristics of the scene.



\begin{figure*}[t]
    \centering
    \includegraphics[width=1\linewidth]{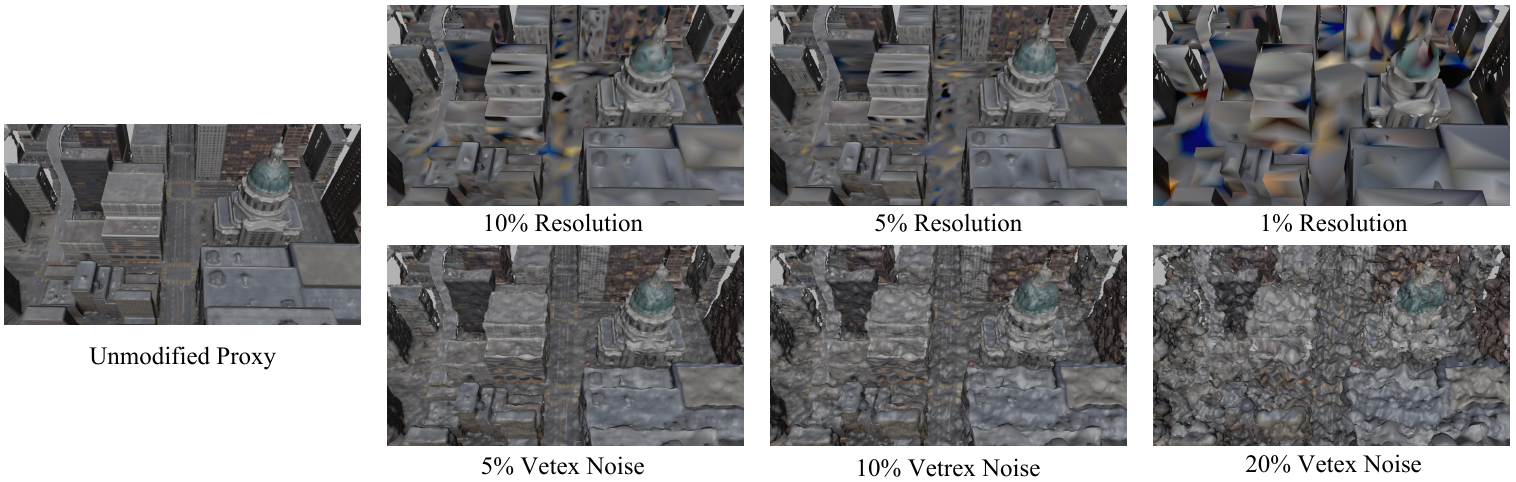}
    \caption{
    Quantitative mesh visualization on different Resolutions and Vertex noise on Block 5.
    }
    \label{fig:ablation_mesh}
\end{figure*}



\subsection{Ablations}
\paragraph{Effect of training procedure.}
As shown in Tab.~\ref{tab:progressive}, we conduct ablation studies on different training strategies. 
\textbf{ID 1} corresponds to the default Octree-GS training and testing pipeline, which serves as our baseline.  
\textbf{ID 2} applies our proxy-guided rendering strategy at test time only, without modifying the Octree-GS training process. Although this setting brings more than a $3\times$ FPS increase, the inconsistency between anchors and their associated Gaussians during training leads to a noticeable drop in rendering quality.

\textbf{ID 3} further enforces consistency by employing proxy-guided rendering also during training. In this case, rendering quality surpasses the baseline, while FPS slightly decreases compared to \textbf{ID~2}, mainly because more anchors grow before being culled by occlusion.
\textbf{ID 4} incorporates the proposed proxy-guided densification strategy and proxy-guided training and rendering. This setting achieves the best balance, delivering further improvements in rendering quality while maintaining a comparable FPS to \textbf{ID~3}.



\begin{table}[h]
\centering
\small
\caption{\textbf{Ablations of different training and inference strategies on Block 5.} {Average anchor} denotes the average number of decoded anchors in the scene.}
\setlength{\tabcolsep}{4pt}
\resizebox{\columnwidth}{!}{
\begin{tabular}{ccccccc}
\toprule[1.1pt]
\multirow{2}{*}{ID} &
  \multirow{2}{*}{\begin{tabular}[c]{@{}c@{}}Occlusion \\ Training \end{tabular}} &
  \multirow{2}{*}{\begin{tabular}[c]{@{}c@{}}Proxy-guided \\ Densification\end{tabular}} &
  \multirow{2}{*}{\begin{tabular}[c]{@{}c@{}}Proxy-guided \\ Inference \end{tabular}} &
  \multirow{2}{*}{\begin{tabular}[c]{@{}c@{}}PSNR$\uparrow$\end{tabular}} &
  \multirow{2}{*}{\begin{tabular}[c]{@{}c@{}}FPS$\uparrow$\end{tabular}} &
  \multirow{2}{*}{\begin{tabular}[c]{@{}c@{}}Average \\ anchor \end{tabular}} \\
     \\
\midrule
1 & {\color[HTML]{CB0000}\XSolidBrush} & {\color[HTML]{CB0000}\XSolidBrush}  & {\color[HTML]{CB0000}\XSolidBrush}  &21.41 & 48 & 719k\\
2 & {\color[HTML]{CB0000}\XSolidBrush} & {\color[HTML]{CB0000}\XSolidBrush}  &  \color[HTML]{009901}\textbf{\Checkmark}  &19.06 & 165 & 82k \\
3 & \color[HTML]{009901}\textbf{\Checkmark} & {\color[HTML]{CB0000}\XSolidBrush}  & \color[HTML]{009901}\textbf{\Checkmark} &21.50 & 147 & 93k\\
4 & \color[HTML]{009901}\textbf{\Checkmark} & \color[HTML]{009901}\textbf{\Checkmark}   & \color[HTML]{009901}\textbf{\Checkmark}  &21.68 &143 &106k \\
\bottomrule[1.1pt]
\end{tabular}
}
\label{tab:progressive}
\end{table}
\vspace{-0.1in}
\paragraph{Rendering time analysis.} In Fig.~\ref{fig:time_comparison}, we quantify the proportion of inference time spent on each component. The lightweight proxy-based depth rendering takes nearly negligible time (around \SI{1}{ms}). Our anchor filtering is also faster due to the reduced number of anchors. The rendering stage is where most of the savings come from: with fewer anchors, both the decoding overhead and Gaussian rasterization are significantly reduced. For more details, we also record the average decode anchors in the Appendix~\ref{Average_anchor_record}.

\begin{table}[t]
\centering
\small 
\caption{\textbf{Integration with different 3DGS rendering accelerations.}
We evaluate our method combined with existing approaches~\citep{Feng2024FlashGSE3,fast_gauss} on Block~1.}
\label{tab:different_rendered}
\setlength{\tabcolsep}{4pt}
\renewcommand{\arraystretch}{1.1}
\begin{tabular*}{\columnwidth}{@{\extracolsep{\fill}}lcccc@{}}
\toprule
Method & PSNR$\uparrow$ & SSIM$\uparrow$ & LPIPS$\downarrow$ & FPS$\uparrow$ \\
\midrule
Original 3DGS & 23.27 & 0.786 & 0.322 & 112 \\
FlashGS       & 23.27 & 0.785 & 0.322 & 115 \\
Hardware 3DGS & 23.20 & 0.781 & 0.328 & 155 \\
\bottomrule
\end{tabular*}
\end{table}
\vspace{-0.1in}
\paragraph{Integration with different 3DGS renderers.}
Since our method primarily optimizes anchors and thus indirectly reduces the number of rendered Gaussians, it can be naturally combined with existing acceleration techniques for the original 3DGS to achieve even higher speed. In Table~\ref{tab:different_rendered}, we evaluate on Block~1. Here, Original 3DGS denotes the default renderer used in Proxy-GS. Replacing it with FlashGS brings a minor improvement, while using a hardware rasterizer for 3DGS slightly compromises rendering quality but further boosts the frame rate by nearly 40 FPS. Note that we reported results with original 3DGS renderer in Tab.~\ref{tab:avg_blocks},~\ref{tab:real_world},~\ref{tab:progressive}. In the Appendix ~\ref{hardware_report}, we report all the results with the hardware 3DGS as the default renderer


\begin{figure}[t]
  \centering
  \includegraphics[width=\columnwidth]{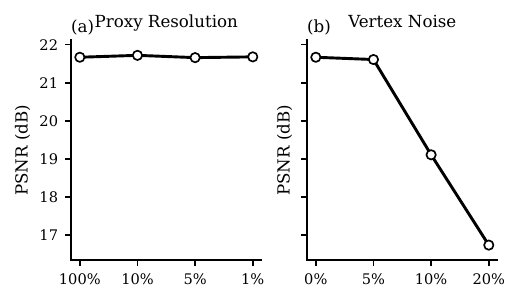}
  \caption{Ablation on proxy fidelity: (a) PSNR vs. Proxy Resolution; (b) PSNR vs. Vertex Noise.}
  \label{fig:proxy_ablation}
  \vspace{-0.1in}
\end{figure}
\vspace{-0.2in}
\paragraph{Dependency on different Proxy Quality.}
To quantify our method’s dependence on the proxy’s accuracy, we conduct two ablations: (i) mesh resolution, we evaluate proxies at multiple resolutions (from fine (\textbf{108MB}) to coarse (\textbf{824KB})) ); and (ii) vertex noise, we perturb mesh vertices with random noise of varying magnitudes. As shown in Fig.~\ref{fig:proxy_ablation}, varying the proxy resolution has only a marginal effect on rendering quality. This is largely because modern urban scenes (buildings, facades, and roads) are dominated by broad, near-planar surfaces, so even coarse proxies preserve the visibility structure needed by our filter. The mesh visualizations in Fig. ~\ref{fig:ablation_mesh} illustrate this effect clearly. When the mesh resolution is reduced, the overall occlusion structure remains correct, only fine details are lost, so the impact on rendering quality is minimal. In contrast, adding vertex noise disrupts the global geometry and breaks the occlusion boundaries, which inherently leads to a much larger degradation in rendering quality. However, As shown in Fig.~\ref{fig:framework}, because there is an inherent offset between the anchors and the decoded Gaussians, our method retains robustness under small perturbations: with noise levels within 5\%, the overall impact remains limited.


%% file: sec/6_conclusion.tex
\section{Conclusion}
In this work, we propose Proxy-GS, a proxy-guided training and inference framework for MLP-based 3D Gaussian Splatting. Our carefully designed proxy-guided filter enables nearly lossless depth acquisition and occlusion culling, while the proxy-guided densification effectively leverages geometric priors from proxies to provide a more structured densification mechanism. Extensive experiments demonstrate that our framework consistently improves both rendering quality and efficiency across diverse scenarios. In particular, on occlusion-rich scenes, Proxy-GS achieves up to a 2.5$\times$ speedup, significantly advancing the practicality of MLP-based methods for VR/AR applications, and establishing a new state-of-the-art in efficient 3D scene representation.

%% file: sec/7_acnowledgement.tex
\section{Acknowledgment}
This research was supported by the Shanghai AI Laboratory and the National Natural Science Foundation of China under Grants 62293543 and 62322605.

%% file: sec/X_suppl.tex
\clearpage
\setcounter{page}{1}
\maketitlesupplementary

\section{Appendix}

\subsection{Division detail on MatrixCity}
\label{partition}
We divide all the Horizon street scenes in MatrixCity's small city into five blocks
(eg. Block 1, Block 2), The partition margin details is in Tab. ~\ref{tab:blocks_magin}
\subsection{Combine with Hardware 3DGS}
\label{hardware_report}
We combine our method with Hardware 3DGS~\citep{fast_gauss} in Tab.~\ref{tab:avg_blocks_hard} and Tab.~\ref{tab:real_world_hard}. As observed, the FPS improves across all datasets, but due to the precision settings used, there is a noticeable decline in rendering quality.

\begin{table}[t]
\centering
\footnotesize
\caption{Partition information in MatrixCity.}
\label{tab:blocks_magin}
\resizebox{\columnwidth}{!}{
\begin{tabular}{c|cccc}
\toprule
Block & $x_{\min}$ & $x_{\max}$ & $y_{\min}$ & $y_{\max}$ \\
\midrule
1 & $-9.80$ & $-2.64$ & $0$ & $3.9$ \\
2 & $-2.64$ & $0.44$ & $0$ & $3.9$ \\
3 & $0.44$ & $3.52$ & $0$ & $3.9$ \\
4 & $3.52$ & $8.70$ & $0$ & $3.9$ \\
5 & $-6.90$ & $6.90$ & $3.9$ & $7.4$ \\
\bottomrule
\end{tabular}
}
\vspace{2pt}
\end{table}

\subsection{Comparison with different depth-acquisition approaches}
In Tab.~\ref{tab:depth-acquisition}, To demonstrate the effectiveness of our depth acquisition strategy, we also evaluate two alternative approaches: directly rendering depth using nvdiffrast~\cite{Laine2020diffrast}, and extracting depth from a pre-trained 3DGS model. These two depth-generation baselines allow us to compare our method against commonly used depth sources.

\begin{table}[t]
\centering
\small 
\caption{\textbf{The FPS of different depth acquisition methods on Block 5.}}
\label{tab:depth-acquisition}
\setlength{\tabcolsep}{4pt}
\renewcommand{\arraystretch}{1.1}

\begin{tabular*}{\columnwidth}{@{\extracolsep{\fill}}cccc@{}}
\toprule
Method & Nvdiffrast & 3DGS & Ours\\
\midrule
FPS & 32 & 54 & 151  \\

\bottomrule
\end{tabular*}
\end{table}

\begin{table*}[h]
\centering
\small
\caption{Combine with Hardware 3DGS ~\citep{fast_gauss}, quantitative results on MatrixCity~\citep{li2023matrixcity}}. 
\label{tab:avg_blocks_hard}
\setlength{\tabcolsep}{4pt}
\renewcommand{\arraystretch}{1.1}
\resizebox{\textwidth}{!}{
\begin{tabular}{l|cccc|cccc|cccc}
\toprule
& \multicolumn{4}{c|}{Block 1\&2} & \multicolumn{4}{c|}{Block 3\&4} & \multicolumn{4}{c}{Block 5} \\
Methods & PSNR$\uparrow$ & SSIM$\uparrow$ & LPIPS$\downarrow$ & FPS$\uparrow$
        & PSNR$\uparrow$ & SSIM$\uparrow$ & LPIPS$\downarrow$ & FPS$\uparrow$
        & PSNR$\uparrow$ & SSIM$\uparrow$ & LPIPS$\downarrow$ & FPS$\uparrow$ \\
\midrule
Proxy-GS & \textbf{22.11} & \textbf{0.751} & \textbf{0.330} & \textbf{126} & \textbf{21.06} & \textbf{0.751} & \textbf{0.348} & \textbf{134} & \textbf{21.68} & \textbf{0.744} & \textbf{0.362} & \textbf{151} \\
+Hardware 3DGS ~\citep{fast_gauss} &\textbf{22.05} & \textbf{0.747} & \textbf{0.338} & \textbf{167} & \textbf{20.86} & \textbf{0.743} & \textbf{0.357} & \textbf{174} & \textbf{21.58} & \textbf{0.735} & \textbf{0.372} & \textbf{196} \\
\bottomrule
\end{tabular}
}
\end{table*}

\begin{table*}[h]
\centering
\small
\vspace{6pt}
\caption{Combine with Hardware 3DGS ~\citep{fast_gauss}, quantitative results on real world Outdoor and Indoor datasets~\citep{xiong2024gauu,kerbl2024hierarchical,barron2023zip}.}
\label{tab:real_world_hard}
\setlength{\tabcolsep}{4pt}
\renewcommand{\arraystretch}{1.1}
\resizebox{\textwidth}{!}{
\begin{tabular}{l|cccc|cccc|cccc}
\toprule
& \multicolumn{4}{c|}{CUHK-LOWER} & \multicolumn{4}{c|}{Berlin} & \multicolumn{4}{c}{Small City} \\
Methods & PSNR$\uparrow$ & SSIM$\uparrow$ & LPIPS$\downarrow$ & FPS$\uparrow$
        & PSNR$\uparrow$ & SSIM$\uparrow$ & LPIPS$\downarrow$ & FPS$\uparrow$
        & PSNR$\uparrow$ & SSIM$\uparrow$ & LPIPS$\downarrow$ & FPS$\uparrow$ \\
\midrule
Proxy-GS &26.44  &0.795  &0.262  &239  &27.85  &0.912  &0.216  &275  &23.09  &0.736  &0.344  &139 \\
+Hardware 3DGS ~\citep{fast_gauss} &26.28 & 0.787 &0.265 & 280 & 27.78 & 0.906 & 0.210 & 325 & 22.91 & 0.732 & 0.343 & 163 \\
\bottomrule
\end{tabular}
}
\end{table*}

\begin{table*}[t]
    \centering
    \caption{Average anchor number used to decode all the datasets}
    \label{tab:avg_anchor_number}
    \begin{tabular}{lcccccc}
        \toprule
        Method & Block 1\&2 & Block 3\&4 & Block 5 & Berlin & CUHK-LOWER & Small City \\
        \midrule
        Proxy-GS  & 190k  & 190k  & 80k  & 40k  & 110k  & 350k \\
        Octree-GS & 800k  & 1040k & 720k & 60k  & 120k  & 840k \\
        \bottomrule
    \end{tabular}
\end{table*}
\subsection{Average decoded anchor number on all the datasets}.
\label{Average_anchor_record}
In Tab.~\ref{tab:avg_anchor_number}, we report the average number of anchors used during training and inference across all datasets. It can be observed that our method consistently reduces the decoding burden, although the degree of improvement varies across different scenes.

\begin{figure}[t]
    \centering
    \includegraphics[width=\linewidth]{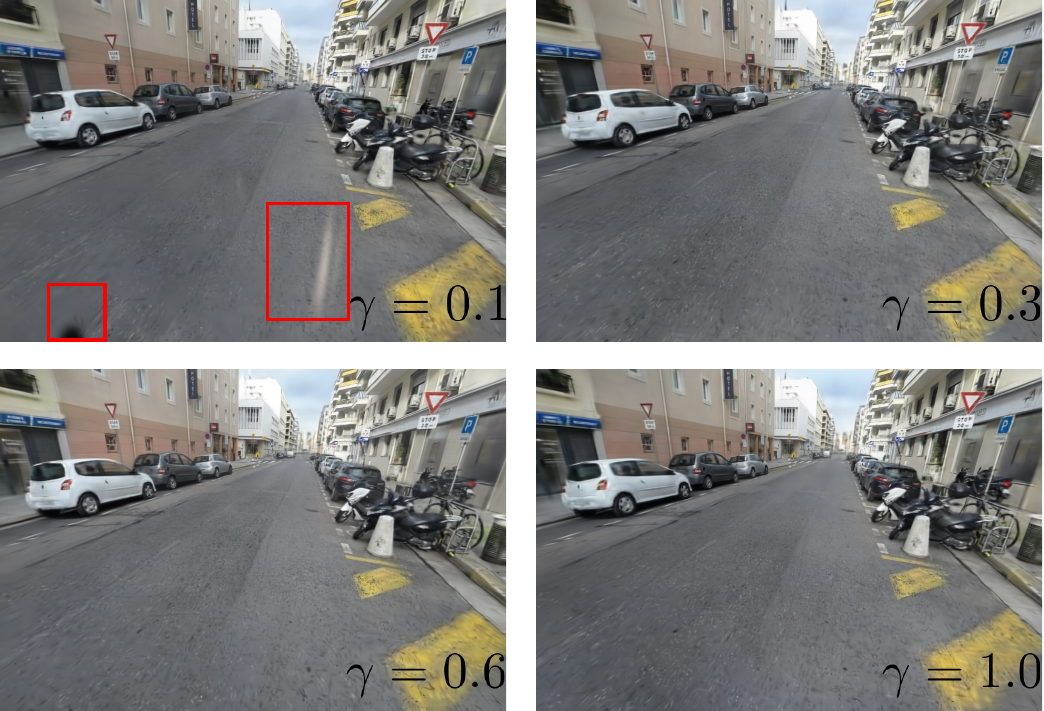}
    \captionsetup{skip=2pt} 
    \caption{\textbf{Visualization on different safety margins.}}
    \label{fig:diffent_gamma}
\end{figure}

\begin{table}[t]
\centering
\small 
\caption{\textbf{Ablations of different safety margin of depth culling $\gamma$
trained on Small City~\cite{kerbl2024hierarchical}.}}
\label{tab:tao}
\setlength{\tabcolsep}{4pt}
\renewcommand{\arraystretch}{1.1}

\begin{tabular*}{\columnwidth}{@{\extracolsep{\fill}}ccccc@{}}
\toprule
$\gamma$ & PSNR$\uparrow$ & SSIM$\uparrow$ & LPIPS$\downarrow$ & FPS$\uparrow$ \\
\midrule
0.1 & 22.94 & 0.734 & 0.349 & 142 \\
0.3 & 23.09 & 0.736 & 0.344 & 139 \\
0.6 & 23.02 & 0.735 & 0.348 & 135 \\
1.0 & 23.05 & 0.736 & 0.345 & 128 \\
\bottomrule
\end{tabular*}
\end{table}

\paragraph{Safety margin of the occlusion culling.}
In Tab.~\ref{tab:tao}, we report results on the Small City dataset by varying the depth culling threshold $\gamma$ in Eq.~\ref{eq:depth_offset}. We observe that $\gamma=0.3$ yields the best trade-off between rendering quality and speed.
As can be seen in Fig.~\ref{fig:diffent_gamma},
when the threshold is too small $\gamma=0.1$, it leads to rendering artifacts in nearby regions. However, setting $\gamma$ too large is also undesirable: a larger threshold introduces excessive anchors, which increases structural redundancy and reduces FPS, while a too small threshold restricts anchor growth and degrades rendering quality.

\begin{figure*}[t]
    \centering
    \includegraphics[width=1\linewidth]{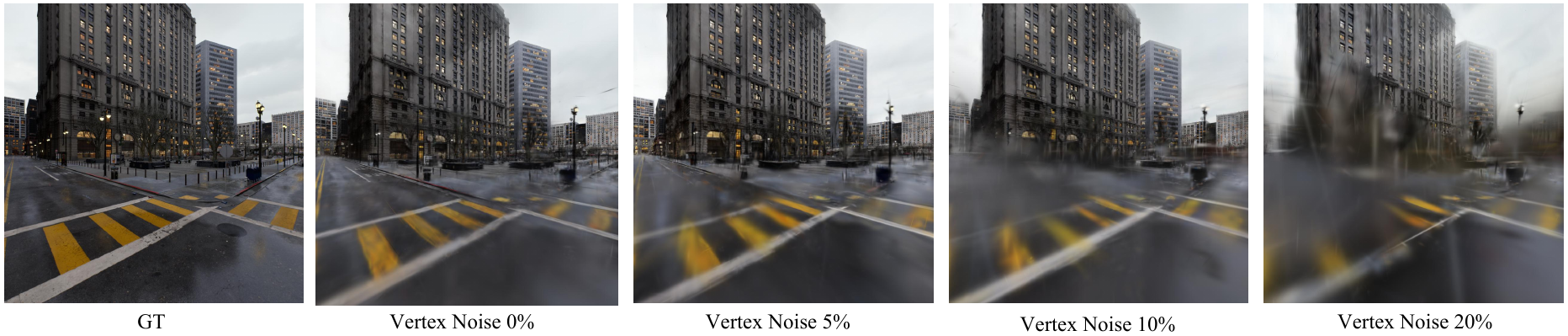}
    \caption{
    Quantitative visualization of different Vertex noise on Block 5.
    }
    \label{fig:ablation_noise}
\end{figure*}

\subsection{Visualization on different Vertex Noise}
As illustrated in Fig.~\ref{fig:ablation_noise}, increasing noise introduces spurious protrusions on the ground, which smear the image and progressively degrade rendering quality.

\subsection{Mesh extraction on different datasets}
\label{mesh_extract}
\subsubsection{Indoor and Outdoor Scenes with Dense Point Clouds}
We describe the mesh extraction process when dense point clouds are available for both indoor and outdoor environments. 
This category includes real-world datasets that provide LiDAR point clouds (e.g.,~\cite{xiong2024gauu}), where mesh generation can be directly performed using surface reconstruction methods, such as ~\citep{huang2023nksr}. 
In addition, for synthetic datasets such as MatrixCity, ground-truth depth maps are available, which can be fused via TSDF to obtain high-quality meshes. 

\subsubsection{Indoor Scenes with Sparse COLMAP Point Clouds}
We describe the workflow of mesh extraction in indoor scenes where only sparse COLMAP reconstructions are available. 
Directly relying on COLMAP to generate dense point clouds in indoor environments is often unreliable, as such scenes frequently contain large textureless regions. 
To address this challenge, To address this challenge, we leverage a foundation-model approach similar to VGGT~\cite{wang2025vggt}. Our experiments show that VGGT-style models exhibit strong robustness in indoor scenes. However, since our pipeline requires alignment with COLMAP poses, we adopt MapAnything~\cite{keetha2025mapanything} instead, using RGB images together with COLMAP inputs to obtain a dense point cloud.
This hybrid strategy enables the recovery of reasonable indoor meshes despite the limitations of sparse COLMAP input.

\subsubsection{Outdoor Scenes with Sparse COLMAP Point Clouds}
For outdoor environments where the reconstruction relies solely on sparse COLMAP point clouds, the abundance of feature points generally mitigates the issue of sparse textures. However, due to the large spatial extent, many 3DGS-based indoor reconstruction methods encounter out-of-memory (OOM) problems when applied to outdoor scenes. To address this, we employ CityGS-X~\citep{gao2025citygs}, a state-of-the-art large-scale geometric reconstruction framework, which leverages multi-GPU parallelism to achieve scalable mesh generation with competitive performance.

\subsubsection{Mesh visualization}
\begin{figure*}[t]
    \centering
    \includegraphics[width=1\linewidth]{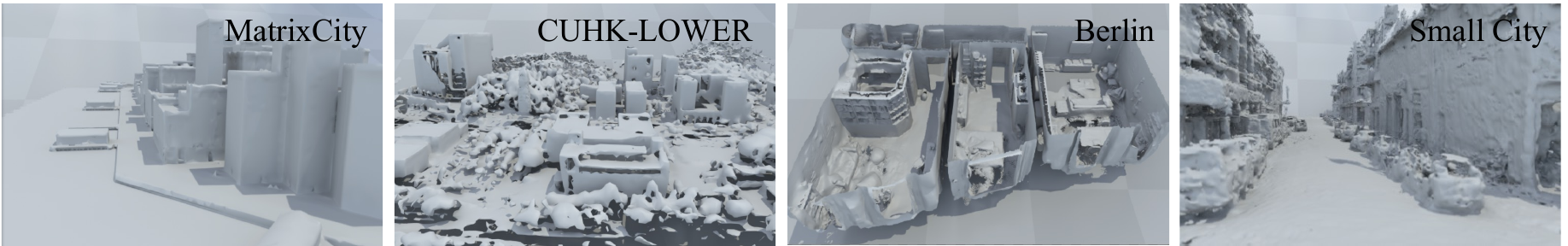}
    \caption{
    \textbf{Mesh visualization}. Scenes include different datasets~\citep{li2023matrixcity, xiong2024gauu, barron2023zip, kerbl2024hierarchical}.
    }
    \label{fig:mesh_visualization}
\end{figure*}

As shown in Fig.~\ref{fig:mesh_visualization}, we visualize all the lightweight proxies. Our method does not require highly accurate meshes; an approximate geometry is sufficient. Thanks to the anchor-based filtering, the subsequent growth of Gaussians introduces offsets that provide additional tolerance, thereby ensuring that our approach maintains a certain degree of robustness to mesh inaccuracies.

\subsection{Fast Depth Acquisition}
\label{sec:fast_depth}

\subsubsection{Overview.}
We follow a modern real-time rendering pipeline to obtain high-quality depth maps at minimal latency. The key ideas are: (i) \emph{preprocess} the reconstructed mesh into compact \emph{clusters}; (ii) perform fully GPU-resident \emph{frustum} and \emph{hierarchical-Z (Hi-Z)} occlusion culling at \emph{cluster} granularity each frame; (iii) emit a \emph{depth-only} pass that leverages Early-Z; and (iv) \emph{zero-copy} the resulting depth buffer into the learning runtime (PyTorch) via Vulkan--CUDA interop, avoiding CPU round trips. This section details each component.

\subsubsection{Preprocessing: from reconstructed mesh to clusters.}
Given a triangle mesh $\mathcal{M}=(\mathcal{V},\mathcal{F})$ obtained by the reconstruction routine above, we apply the following:
\begin{enumerate}
\item \textbf{Topology-preserving simplification.}
We reduce face count with a quadric-error-metric (QEM) style simplifier while enforcing feature and boundary preservation.
For a vertex in homogeneous coordinates $\tilde{\mathbf{x}}=(x,y,z,1)^\top$ and its incident face planes
$\{\mathbf{p}_f=(a,b,c,d)^\top\}$ (with $\|(a,b,c)\|_2=1$ for all $f$), the local quadric is
\[
Q \;=\; \sum_{f} \mathbf{p}_f \mathbf{p}_f^\top,\qquad
\]
These per-vertex quadrics are \emph{accumulated} and then used by an edge-collapse procedure to decide the contraction position and cost, which removes superfluous micro-triangles commonly produced by reconstruction and improves cache locality and GPU occupancy.

\paragraph{Edge-collapse simplification with QEM.}
For each vertex $v$, accumulate $Q_v=\sum_{f\in N(v)}\mathbf{p}_f\mathbf{p}_f^\top$.
To collapse an edge $(i,j)$, combine quadrics
\[
Q' \;=\; Q_i + Q_j,\qquad E(\tilde{\mathbf x}) \;=\; \tilde{\mathbf x}^\top Q' \tilde{\mathbf x}.
\]
Partition $Q'$ as $Q'=\begin{bmatrix}A & \mathbf b\\ \mathbf b^\top & c\end{bmatrix}$ with
$A\!\in\!\mathbb{R}^{3\times3}$, $\mathbf b\!\in\!\mathbb{R}^3$, $c\!\in\!\mathbb{R}$.
The optimal contraction position is
\begin{equation}
\begin{aligned}
\mathbf{x}^{*} &= \operatorname*{arg\,min}_{\mathbf{x}\in \mathbb{R}^3}
\; \mathbf{x}^\top A\,\mathbf{x} + 2\,\mathbf{b}^\top \mathbf{x} + c \\
&= -\,A^{-1}\mathbf{b} \quad \text{(if $A$ is invertible).}
\end{aligned}
\end{equation}

with cost $\delta = E([\mathbf{x}^{*\top},\,1]^\top)$.

If $A$ is singular, evaluate $\{\mathbf x_i,\mathbf x_j,(\mathbf x_i+\mathbf x_j)/2\}$ and pick the one with minimal $E$.
We maintain a priority queue keyed by $\delta$ and iteratively collapse the lowest-cost edge, updating connectivity and setting the new vertex quadric to $Q'$. Collapses that would break manifoldness or flip triangle orientations are forbidden.

\paragraph{Boundary/feature preservation.}
For a boundary or sharp-crease edge with unit tangent $\mathbf t$ and unit average normal $\hat{\mathbf n}$,
add two \emph{constraint planes} whose intersection is the edge line,
\[
\mathbf p_1 = (\hat{\mathbf n},\ -\hat{\mathbf n}^\top \mathbf x_0)^\top,\quad
\mathbf p_2 = (\widehat{\mathbf t\times \hat{\mathbf n}},\ -\,\widehat{\mathbf t\times \hat{\mathbf n}}^{\!\top}\mathbf x_0)^\top,
\]
and augment incident vertex quadrics by
\[
Q_v \leftarrow Q_v + \lambda_b\,\mathbf p_1\mathbf p_1^\top + \lambda_b\,\mathbf p_2\mathbf p_2^\top,
\]
with a large weight $\lambda_b$. Alternatively, restrict collapses so boundary vertices only collapse along the boundary,
and forbid collapses across edges whose dihedral angle exceeds a feature threshold.

\item \textbf{Cluster construction.}
We partition the simplified mesh into triangle sets $\{\mathcal{L}_k\}_{k=1}^{K}$ such that $\bigsqcup_{k}\mathcal{L}_k=\mathcal{F}$ and $\tau_{\min}\!\le\!|\mathcal{L}_k|\!\le\!\tau_{\max}$. For each cluster we precompute:
(a) an object-space axis-aligned bounding box (AABB) $\mathrm{AABB}_k=[\mathbf{b}_k^{\min},\mathbf{b}_k^{\max}]$; and
(b) a conservative screen-space bounding rectangle at level-0, $R_k^{(0)}$, for any given view.
Project the AABB's eight corners $\{\mathbf{x}_{k,j}\}_{j=1}^{8}$ with the view-projection $PV$:
\[
\mathbf{y}_{k,j}=PV\!\begin{bmatrix}\mathbf{x}_{k,j}\\1\end{bmatrix},\qquad
\mathbf{u}_{k,j}^{\text{ndc}}=\Big(\tfrac{y^x_{k,j}}{y^w_{k,j}},\,\tfrac{y^y_{k,j}}{y^w_{k,j}}\Big).
\]
Let the viewport be $W\times H$ (origin at the top-left). Map to pixels
\[
\mathbf{s}_{k,j}=\Big(\tfrac{W}{2}(\,u^{\text{ndc}}_{x}+1),\ \tfrac{H}{2}(u^{\text{ndc}}_{y} + 1)\Big),
\]
then take an outward-rounded, padded box (padding $\Delta\!\in\!\{0,1\}$) and clip to the screen:
\begin{equation}
\begin{aligned}
R_k^{(0)}
&= \Bigl[\, \lfloor \min_j \mathbf{s}_{k,j} \rfloor - \Delta,\ \ \lceil \max_j \mathbf{s}_{k,j} \rceil + \Delta \,\Bigr] \\
&\qquad \cap\ [0, W\!-\!1]\times [0, H\!-\!1].
\end{aligned}
\end{equation}

Such cluster construction helps us to do cluster-level culling, increasing granularity compared to per-triangle culling while retaining high selectivity.

\end{enumerate}

\paragraph{Per-frame visibility: frustum and Hi-Z occlusion.}
Let $\{\Pi_i\}_{i=1}^6$ be the frustum planes with \emph{inward} normals $\mathbf n_i$ and offsets $d_i$.
A cluster $\mathcal L_k$ with $\mathrm{AABB}_k$ corners $\{\mathbf x_j\}_{j=1}^8$ is frustum-culled if
\begin{equation}
\exists\, i\ \text{s.t.}\ \max_{j}\bigl(\mathbf n_i^\top \mathbf x_j + d_i\bigr) < 0 .
\label{eq:frustum}
\end{equation}
Let $Z^{(0)}(u,v)$ be the base depth. The Hi-Z pyramid for standard depth is
\begin{equation}
Z^{(\ell+1)}(u,v)=\max_{\delta_x,\delta_y\in\{0,1\}}
Z^{(\ell)}(2u+\delta_x,\,2v+\delta_y).
\label{eq:hiz_build}
\end{equation}

\paragraph{Level snapping and conservative depth.}
Given $R_k^{(0)}$, choose a pyramid level $\ell$ (e.g., 
$\ell=\mathrm{clamp}(\lfloor\log_2(\max(\mathrm{width}(R_k^{(0)}),\mathrm{height}(R_k^{(0)})))\rfloor-c,\ 0,\ L_{\max})$
with a small constant $c\!\in\!\{1,2\}$), and snap the rectangle to level $\ell$ by outward rounding:
\[
R_k^{(\ell)}=\Big[\ \Big\lfloor \tfrac{R^{(0)}_{k,\min}}{2^{\ell}}\Big\rfloor,\ \Big\lceil \tfrac{R^{(0)}_{k,\max}}{2^{\ell}}\Big\rceil\ \Big].
\]
Let $\mathbf y_{k,j}=PV\!\big[\mathbf x_{k,j}^{\top},\,1\big]^{\top}$ denote the
clip-space 4-vectors of the eight AABB corners introduced above (the same ones
used to build $R_k^{(0)}$). A conservative near-depth estimate for the cluster is
\[
\hat z_k \;=\; \min_{j=1,\ldots,8}\;
\!\left(\max\!\left(z^{\text{ndc}}_{\text{near}},
\ \frac{y^z_{k,j}}{y^w_{k,j}}\right)\right),
\]
If any $y^w_{k,j}\le 0$, the near-plane clamp above makes the estimate
conservative; alternatively, one may skip the occlusion test for full safety.

Given the screen-space bounding box $R_k$ of $\mathcal L_k$ snapped to level $\ell$, and a conservative near depth
$\hat z_k$ of $\mathcal L_k$, the occlusion test is
\begin{equation}
\text{occluded}(\mathcal L_k)\iff
\hat z_k \ge \max_{(u,v)\in R_k^{(\ell)}} Z^{(\ell)}(u,v).
\label{eq:hiz_test}
\end{equation}

\paragraph{Depth-only pass with early-Z.}
After visibility, we render only the surviving clusters in a \emph{solid, depth-only} pipeline (color writes disabled, depth writes enabled). 
A minimal fragment shader lets the rasterizer perform early-depth testing. This produces the depth map $D\in\mathbb{R}^{H\times W}$ used downstream.

\paragraph{Zero-copy interop to PyTorch.}
In order to obtain the depth every frame efficiently, a naive path would be to read back the GPU depth buffer to host memory and then upload it to CUDA, introducing synchronization and PCIe traffic. 
Instead, we adopt a fully GPU-resident path: we render with Vulkan and export the depth image’s memory as an \emph{external file descriptor} (FD). On the CUDA side, we import that FD as external memory and map it to a device pointer; the pointer is then wrapped as a PyTorch CUDA tensor without a copy.